\documentclass{article}

\usepackage{arxiv}

\usepackage{amsmath,amsfonts,bm}

\def\eqref#1{equation~\ref{#1}}

\def\1{\bm{1}}

\def\vmu{{\bm{\mu}}}
\def\vtheta{{\bm{\theta}}}

\def\vk{{\bm{k}}}

\def\vs{{\bm{s}}}

\def\vx{{\bm{x}}}
\def\vy{{\bm{y}}}

\def\mI{{\bm{I}}}

\def\mK{{\bm{K}}}

\def\mX{{\bm{X}}}

\DeclareMathAlphabet{\mathsfit}{\encodingdefault}{\sfdefault}{m}{sl}
\SetMathAlphabet{\mathsfit}{bold}{\encodingdefault}{\sfdefault}{bx}{n}

\def\sR{{\mathbb{R}}}

\DeclareMathOperator*{\argmax}{arg\,max}
\DeclareMathOperator*{\argmin}{arg\,min}

\usepackage{amssymb}
\usepackage{algorithm}
\usepackage[noend]{algpseudocode}
\usepackage{enumerate}   

\usepackage[utf8]{inputenc} 
\usepackage[T1]{fontenc}    
\usepackage{hyperref}       
\usepackage{url}            
\usepackage{booktabs}       
\usepackage{amsfonts}       
\usepackage{nicefrac}       
\usepackage{microtype}      
\usepackage{cleveref}       
\usepackage{lipsum}         
\usepackage{graphicx}
\usepackage{natbib}
\usepackage{doi}

\title{Large-Scale Bayesian Optimisation Using Uncertainty-Calibrated Product-of-Experts Gaussian Process Models with Trust Regions}

\date{}

\newif\ifuniqueAffiliation
\uniqueAffiliationtrue

\ifuniqueAffiliation 
\author{ \href{https://orcid.org/0000-0002-8452-0999}{\includegraphics[scale=0.06]{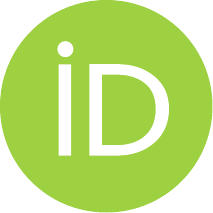}\hspace{1mm} Yean Hoon Ong} \\
	University College London\\
	London, United Kingdom \\
	\texttt{yean-hoon.ong@ucl.ac.uk} \\
}

\renewcommand{\headeright}{}
\renewcommand{\undertitle}{}
\renewcommand{\shorttitle}{Bayesian optimisation using product-of-experts Gaussian process models with trust regions}

\hypersetup{ 
pdftitle={Large-Scale Bayesian Optimisation Using Uncertainty-Calibrated Product-of-Experts Gaussian Process Models with Trust Regions}, 
pdfsubject={cs.LG}, 
pdfauthor={Yean Hoon Ong}, 
pdfkeywords={Bayesian optimisation, Product-of-experts Gaussian processes, Uncertainty calibration, Trust regions, Scalable Gaussian processes}, 
}

\begin{document}
\maketitle

\begin{abstract}
The use of collections of independent local Gaussian processes (GP-ind) combined with trust regions is a widely adopted approach for large-scale and high-dimensional Bayesian optimisation (BO). However, this approach suffers from several limitations, including discontinuities at the boundaries of disjoint regions, an inability to capture long-range spatial correlations, potential information loss, and abrupt increases in computational overhead. 
To address these limitations, this paper introduces a novel BO algorithm, termed BO-pro-c-tr. The proposed method employs a product-of-experts Gaussian process model with uncertainty calibration (GP-pro-c) as its surrogate model and incorporates a one-to-many association strategy for integrating GP-pro-c with trust regions. 
Empirical results suggest that, when combined with the GP-LCB acquisition function, BO-pro-c-tr achieves an average reduction of 2.6\% in simple regret and 34.9\% in computational overhead relative to BO algorithms based on independent local GP models coupled with trust regions.
\end{abstract}


\section{Introduction}
\label{sec:sec_intro}

Bayesian optimisation (BO) \citep{Brochu2010, Shahriari2016p, Frazier2018, garnett_bayesoptbook_2023} is a widely used approach for optimising expensive black-box objective functions. The effectiveness of BO relies heavily on the quality of its surrogate model, which is typically a Gaussian process (GP). By providing both predictive means and uncertainty estimates, GP models enable efficient exploration of the search space while limiting the number of expensive function evaluations. 

As optimisation problems become larger and higher-dimensional, however, conventional GP-based BO faces significant scalability challenges. Most BO methods employ a single global GP surrogate, whose training and prediction procedures require operations on an $n \times n$ covariance matrix, where $n$ denotes the number of observed data points \citep{Rasmussen2006, Shahriari2016p}. The resulting $\mathcal{O}(n^3)$ computational complexity can become prohibitive as the number of evaluations increases.

To alleviate this computational burden, recent research has explored local surrogate models and trust-region methods, which focus computational effort on promising regions of the search space while maintaining scalability in large optimisation problems.
A prominent example is the TuRBO algorithm proposed by \cite{Eriksson2019}. TuRBO employs multiple independent local GP models, each producing its own posterior predictions and acquisition values. It further adopts a one-to-one association between local GP models and trust regions, thereby restricting the search process to promising subregions of the input space rather than the entire domain.

Although effective in many settings, this approach has several limitations. First, the absence of information sharing between local GP models prevents the capture of global, long-range spatial correlations. Second, the repeated shrinking and expansion of trust regions within small regions of the search space can cause optimisation to become trapped in local minima. Third, the number of observations assigned to an individual local GP may grow sufficiently large that the model effectively behaves as a global GP, thereby reintroducing the $\mathcal{O}(n^3)$ computational bottleneck.

To address these limitations, this paper adopts the product-of-experts GP model with uncertainty calibration proposed by \citep{Ong2026}. This model, referred to as GP-pro-c, consists of multiple local GP experts whose predictions are combined through an aggregation mechanism. GP-pro-c offers several advantages: (i) it captures local patterns through region-specific hyperparameter learning; (ii) it preserves global, long-range spatial correlations; (iii) it alleviates the cubic computational complexity associated with GP regression; and (iv) it incorporates an information-based uncertainty calibration mechanism that mitigates variance overestimation in local GP models.

The objective of this paper is to develop a new BO algorithm, termed BO-pro-c-tr, that combines the strengths of GP-pro-c with trust-region methods. The proposed approach addresses several limitations of TuRBO through three key design choices. First, all evaluated observations are retained throughout optimisation, avoiding information loss. Second, a ball-tree assignment strategy is used to construct local GP models while maintaining balanced dataset sizes across experts, thereby preventing abrupt increases in computational overhead. Third, a one-to-many association strategy is introduced, allowing each local GP model to be linked to multiple trust regions and reducing the likelihood of optimisation becoming trapped in local regions of the search space.

The main contributions of this paper are as follows: 
\begin{itemize} 
\item Introduction of a novel Bayesian optimisation algorithm, BO-pro-c-tr, for large-scale and high-dimensional optimisation. 
\item Comprehensive empirical evaluation of BO-pro-c-tr across a diverse range of BO settings using three widely adopted acquisition functions: expected improvement (EI), Gaussian process lower confidence bound (GP-LCB), and Thompson sampling (TS). 
\item Rigorous assessment of optimisation performance and computational overhead, including comparisons with TuRBO and BO methods based on a single global GP model. 
\end{itemize}

The remainder of this paper is organised as follows. 
Section~\ref{sec:sec_s2_related_work} reviews the related literature. 
Section~\ref{sec:sec_c5_algo} presents the BO-pro-c-tr algorithm. 
Section~\ref{sec:sec_c5_experiments} reports the experimental results. 
Finally, Section~\ref{sec:sec_s5_conclusion} concludes the paper.

\section{Related Work}
\label{sec:sec_s2_related_work}

The TuRBO algorithm proposed by \cite{Eriksson2019} employs a collection of independent local GP models, each trained on a disjoint subset of the data and producing its own posterior predictions. Each local GP is associated with a trust region that restricts the search process to a hyper-rectangular neighbourhood centred on the current best observation within that model.

At each optimisation iteration $t$, TuRBO applies Thompson sampling to generate candidate points within individual trust regions, selects a batch of query points from the union of these regions, and updates the corresponding local optimisation problems. The size of each trust region is adjusted according to predefined success and failure criteria. If successive query points sampled from a trust region improve upon previously observed objective values, the region is expanded to encourage exploration. Conversely, after a sufficient number of unsuccessful evaluations, the region is shrunk to promote local exploitation.

TuRBO has demonstrated strong empirical performance on large-scale and high-dimensional BO problems, outperforming several state-of-the-art alternatives, including additive GP models \citep{Eriksson2019}. Nevertheless, several fundamental limitations remain.

First, the independence of local GP models prevents information sharing across regions. Consequently, predictions may become discontinuous at region boundaries, and the model is unable to capture global, long-range spatial correlations \citep{Samo2015, Park2016, Liu2020}. This limitation directly affects the quality of acquisition functions, which rely on coherent uncertainty estimates throughout the search space.

Second, TuRBO sacrifices data efficiency in favour of robustness. When a trust region shrinks below a predefined threshold, the corresponding local GP model is discarded and replaced with a newly initialised model. Although this mechanism helps prevent stagnation in overly local regions, it also results in the permanent loss of previously evaluated observations. In BO settings, where function evaluations are typically expensive, this represents a significant inefficiency.

Third, the scalability benefits of TuRBO are not guaranteed throughout the optimisation process. As optimisation progresses, the number of observations assigned to a local GP model may increase substantially. Before the discard mechanism is activated, a local GP can accumulate enough observations to behave similarly to a global GP, thereby reintroducing the $\mathcal{O}(n^3)$ computational bottleneck. This exposes an inherent trade-off between local modelling and computational scalability that TuRBO does not fully resolve.

An alternative direction was explored by \cite{Tautvaisas2022}, who combined a product-of-experts GP model with trust-region methods. While this approach partially addresses the lack of global consistency associated with independent local GP models, its effectiveness remains limited. The empirical evaluation was restricted to simplified BO settings involving sequential query selection and the UCB acquisition function with a fixed parameter $\kappa = 1$, and the reported performance improvements were modest and inconsistent. The authors attributed these limitations to variance overestimation in product-of-experts GP models but did not propose a principled solution. Instead, fixing $\kappa = 1$ served primarily as a heuristic mechanism for reducing over-exploration rather than addressing the underlying modelling issue.

More recently, \cite{Ong2026} introduced the GP-pro-c model, which directly addresses variance overestimation in product-of-experts GP models through an information-theoretic uncertainty calibration mechanism. By exploiting the monotonicity and submodularity properties of information gain, the method derives a principled scaling factor for calibrating the predictive variance of each local expert. This approach improves uncertainty quantification while maintaining predictive accuracy and reducing computational overhead.

Despite recent progress, a key challenge remains unresolved: developing a scalable Bayesian optimisation framework that combines the efficiency of trust-region methods with the global consistency of collaborative Gaussian process models. Existing approaches do not simultaneously provide information sharing, calibrated uncertainty estimates, data retention, and computational scalability. To the best of the author's knowledge, no previous work has integrated an uncertainty-calibrated product-of-experts GP model with trust-region methods for Bayesian optimisation. 

To address this challenge, this paper proposes BO-pro-c-tr, which combines uncertainty-calibrated product-of-experts GP modelling with a redesigned trust-region mechanism. The proposed framework enables information sharing across local regions, retains all evaluated observations, and maintains balanced computational costs across local GP models.

\section{Algorithm BO-pro-c-tr}
\label{sec:sec_c5_algo}

Algorithm~\ref{alg:gp_poe_c_tr} presents the pseudo-code of BO-pro-c-tr. An implementation of BO-pro-c-tr is available at \url{https://github.com/yhong123/bo-pro-c-tr}. 

Given an unknown noisy objective function $f : \mathcal{X} \rightarrow \sR$, where $\mathcal{X}$ is a compact subset of $\sR^d$, the objective is to identify the minimiser $\vx_*$ within the input domain $\mathcal{X}$ under a fixed evaluation budget $T$.

\begin{algorithm} 
\caption{BO-pro-c-tr} 
\label{alg:gp_poe_c_tr} 
\begin{algorithmic} 
	\State \textbf{Input:} unknown noisy objective function $f$ 
	\State \textbf{Input:} input domain $\mathcal{X}$ 
	\State \textbf{Input:} number of local GP models $M$ (minimum: 2) 
	\State \textbf{Input:} initial sample size $N_{\mathrm{init}}$ 
	\State \textbf{Input:} maximum number of data points per local GP model $N_{\mathrm{GP}}$ 
	\State \textbf{Input:} acquisition function $a$ 
	\State \textbf{Input:} batch size $B$ 
	\State \textbf{Input:} evaluation budget $T$ 
	\State \textbf{Input:} success threshold $\tau_{\mathrm{succ}}$ 
	\State \textbf{Input:} failure threshold $\tau_{\mathrm{fail}}$ 
	\State \textbf{Input:} inactivity threshold $\tau_{\mathrm{inac}}$ 
	\State \textbf{Input:} trust-region side lengths $L_{\mathrm{init}}$, $L_{\mathrm{min}}$, $L_{\mathrm{max}}$ 
	\State \textbf{Initialise:} $\mathcal{D}_0 = \{(\vx_i,y_i)\}_{i \le N_{\mathrm{init}}}$ 
	\For{$t = 1,\ldots,T$} 
		\State $M \gets \max\!\left(2,\mathrm{int}(|\mathcal{D}_{t-1}|/N_{\mathrm{GP}})\right)$ 
		\State Partition $\mathcal{D}_{t-1}$ into $M$ subsets $\mathcal{D}^{(m)}_{t-1}$ 
		\State Assign each subset $\mathcal{D}^{(m)}_{t-1}$ to local expert $GP^{(m)}_{t-1}$ 
		\State Train each $GP^{(m)}_{t-1}$ using $\mathcal{D}^{(m)}_{t-1}$ 
		\For{$m = 1,\ldots,M$} 
			\If{no trust region centred on the best observation of $GP^{(m)}_{t-1}$ exists} 
				\State Create a new trust region 
			\EndIf 
		\EndFor 
		\State Initialise candidate set: $C_t \gets \varnothing$ 
		\For{each trust region $TR^{(q)}$} 
			\State Generate candidate set $C_t^{(q)} = \{\vx'\}$ within $TR^{(q)}$ 
			\State Update candidate set: $C_t \gets C_t \cup C_t^{(q)}$ 
		\EndFor 
		\For{each candidate point $\vx'_i \in C_t$} 
			\State Compute calibration ratio (Eq.~\ref{eq:eq_c6_cc_ratio})
			\State Compute aggregated posterior mean and variance (Eqs.~\ref{eq:eq_c6_agg_mu}--\ref{eq:eq_c6_agg_sigma}) 
		\EndFor 
		\State Construct a batch of query points $\{\vx_{t,j}\}_{j=1}^{B}$ greedily from $C_t$
		    	\For{$j = 1,\ldots,B$}
			        	\State Select query point $\vx_{t,j}$:
				\begin{align}
        				\vx_{t,j} =
        				\begin{cases}
        					\argmax_{\vx' \in C_t}  a^{\mathrm{EI}}  (\vx' \mid \mathcal{D}_{t-1}),  \\[4pt]
        					\argmin_{\vx' \in C_t} a^{\mathrm{LCB}}  (\vx' \mid \mathcal{D}_{t-1}),  \\[4pt]
        					\argmin_{\vx' \in C_t}  a^{\mathrm{TS}}   (\vx' \mid \mathcal{D}_{t-1}),
        				\end{cases}
        				\end{align}
        				\State Remove $\vx_{t,j}$ from candidate set $C_t$
    			\EndFor

    		\For{$j = 1,\ldots,B$}
        			\State Obtain observation    $y_{t,j} = f(\vx_{t,j}) + \epsilon_{t,j}$
   		\EndFor
		\State Update dataset: $ \mathcal{D}_t \gets \mathcal{D}_{t-1} \cup \{(\vx_{t,j},y_{t,j})\}_{j \le B} $
		\For{each query point $\vx_{t,j}$} 
			\If{$\vx_{t,j}$ is associated with $TR^{(q)}$} 
				\State Update the side length of $TR^{(q)}$ 
			\EndIf 
		\EndFor 
		\For{each trust region $TR^{(q)}$} 
			\If{$TR^{(q)}$ satisfies the discard criterion} 
				\State Discard $TR^{(q)}$ 
			\EndIf 
		\EndFor 
	\EndFor 
	\State \textbf{return} $ \hat{\vx}_T = \argmin_{\vx \in \mathcal{D}_T} y(\vx) $
\end{algorithmic}
\end{algorithm}

\paragraph{Initialisation} \hfill 

The algorithm begins by constructing an initial dataset $\mathcal{D}_{0} = \{(\vx_i, y_i)\}_{i \leqslant N_{init}}$, where $y_i = f(\vx_i) + \epsilon_i$. The input locations $\vx_i$ are generated using Latin hypercube sampling \citep{Jones1998, Bull2011}. Following common practice in the BO literature, the initial sample size is set to $N_{init} = 2(d+1)$, where $d$ denotes the input dimensionality \citep{balandat2020botorch}.

\paragraph{Data assignment and local GP training} \hfill

Each local GP model is constrained to a maximum of $N_{GP} \in \{200, 400\}$ data points. At iteration $t$, given dataset $\mathcal{D}_{t-1}$ containing $n$ observations, the number of local GP models is defined as \[ M = \begin{cases} 2, & |\mathcal{D}_{t-1}| < 2N_{GP},\\ \mathrm{int}(|\mathcal{D}_{t-1}|/N_{GP}), & \text{otherwise}. \end{cases} \]
This strategy ensures that each partition contains at most $N_{GP}$ data points while maintaining balanced workloads across local models.

The dataset is partitioned into subsets $\mathcal{D}_{t-1}^{(m)}$ using the ball-tree assignment method (see \cite{Ong2026} for the details this assignment method). This partitioning strategy maintains balanced data allocations and prevents any individual GP model from becoming computationally expensive.

Each local model $GP_{t-1}^{(m)}$ is trained using a constant mean function and a Matérn-$5/2$ kernel with automatic relevance determination (ARD). The hyperparameters $\vtheta_{t-1}^{(m)}$ are estimated by maximising the log marginal likelihood:
\begin{align}
\log p \left( \vy_{t-1}^{(m)} \mid \mX_{t-1}^{(m)},  \vtheta_{t-1}^{(m)} \right) = &- \frac{1}{2} \left( \vy_{t-1}^{(m)} - \tilde{\vmu}_{t-1}^{(m)} \right)^{\top} \left( \mK_{t-1}^{(m)} + \ddot{\sigma}_{t-1,m}^2 \ \mI \right)^{-1} \left( \vy_{t-1}^{(m)} -  \tilde{\vmu}_{t-1}^{(m)} \right)  \notag \\
								    &-  \frac{1}{2} \log \left| \mK_{t-1}^{(m)} + \ddot{\sigma}_{t-1,m}^2 \ \mI \right| - \frac{ \left| \mathcal{D}_{t-1}^{(m)} \right| }{2} \log 2 \pi   
\end{align}
where $\tilde{\vmu}_{t-1}^{(m)}$ denotes the vector of mean-function values evaluated at $\mX_{t-1}^{(m)}$, $\ddot{\sigma}_{t-1,m}^{2}$ denotes the observation noise variance, and $\mK_{t-1}^{(m)} + \ddot{\sigma}_{t-1,m}^{2}\mI$ is the covariance matrix associated with the noisy observations $\vy_{t-1}^{(m)}$.

\paragraph{Creating and initialising trust regions} \hfill

Trust regions restrict candidate generation to promising regions of the search space, thereby avoiding unnecessary exploration of the entire input domain. At each optimisation iteration, a set of $Q$ trust regions is maintained, consisting of both newly created regions and active regions carried over from previous iterations.

Each trust region $TR^{(q)}$ is represented by a hyper-rectangle characterised by: 
\begin{itemize} 
\item a centre location; 
\item a side length for each input dimension; 
\item a success counter, recording the number of times candidates drawn from $TR^{(q)}$ improve upon the current best observation; \item a failure counter, recording the number of times all candidates drawn from $TR^{(q)}$ fail to improve upon the current best observation; and 
\item an inactivity counter, recording the number of optimisation iterations in which no candidate from $TR^{(q)}$ is selected for evaluation. 
\end{itemize}

The trust-region hyperparameters are set to \[ \tau_{succ}=2,\quad \tau_{fail}=2,\quad \tau_{inac}=4,\quad L_{init}=0.8,\quad L_{min}=2^{-7},\quad L_{max}=1.6 \]
These values assume that the input space has been scaled to the unit hypercube $[0,1]^d$. The parameters $\tau_{succ}$, $\tau_{fail}$, and $\tau_{inac}$ define the thresholds for the success, failure, and inactivity counters, respectively, whereas $L_{init}$, $L_{min}$, and $L_{max}$ denote the initial, minimum, and maximum base side lengths of a trust region.

At iteration $t$, a new trust region $TR^{(q)}$ is created for local model $GP_{t-1}^{(m)}$ if 
(i) no existing trust region is centred on the best observation associated with $GP_{t-1}^{(m)}$, and 
(ii) the maximum side length among all active trust regions is smaller than $L_{init}/2$.
The newly created trust region is initialised with base side length $L \leftarrow L_{init}$, and its centre is placed at the input location corresponding to the best observation of $GP_{t-1}^{(m)}$.

\begin{figure}
\centering
\includegraphics{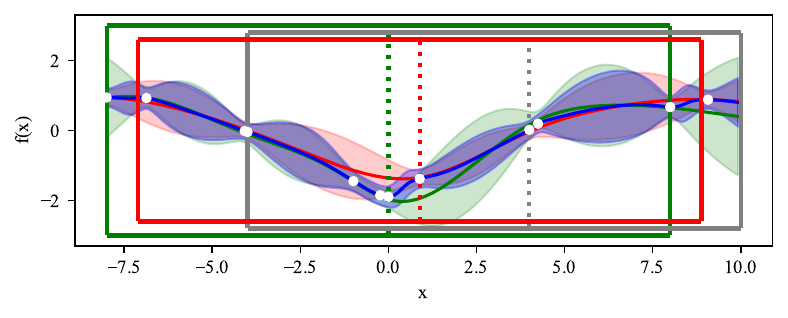}
\caption{ Illustration of the GP-pro-c model with trust regions in BO-pro-c-tr for a one-dimensional optimisation problem. The green and red curves (with shaded uncertainty bands) represent the posterior predictions of two local GP models, while the blue curve (with shaded uncertainty band) denotes the aggregated GP-pro-c prediction. The figure also illustrates the association between two local GP models and three trust regions. The green and red hyper-rectangles represent newly created trust regions centred on the best observations of their respective local models, whereas the grey hyper-rectangle represents an active trust region carried over from a previous optimisation iteration. The dotted vertical lines indicate the centres of the trust regions. }
\label{fig:fig_c5_gpoe_c_tr2}
\end{figure}

Figure~\ref{fig:fig_c5_gpoe_c_tr2} illustrates the association between two local GP models and three trust regions for a one-dimensional optimisation problem. Each trust region is associated with exactly one local GP model, whereas a local GP model may be associated with multiple trust regions. The green and red hyper-rectangles represent newly created trust regions centred on the best observations of their respective local GP models, while the grey hyper-rectangle represents an active trust region carried over from a previous optimisation iteration.

This one-to-many association introduces an important distinction from conventional trust-region BO methods. The association between trust regions and local GP models is flexible rather than permanent. Consequently, when a trust region is discarded, the associated local GP model and its training data are retained. This design eliminates the information loss encountered in methods such as TuRBO, enables efficient reuse of all previously evaluated observations, and preserves the global predictive capabilities of the GP-pro-c model.

\paragraph{Generating candidate points} \hfill

At iteration $t$, each trust region $TR^{(q)}$ generates a candidate set of size \[ \frac{1}{Q}\min(200d, 10^4) \] where $Q$ denotes the total number of trust regions. Each candidate set is generated using a scrambled Sobol sequence within the intersection of the trust region and the domain $[0,1]^d$. Consequently, candidate generation for each $TR^{(q)}$ is restricted to its associated region of the search space.

\paragraph{Aggregating posterior predictions } \hfill 

Let $\mathcal{X}'$ denote the union of all candidate sets generated by the $Q$ trust regions. For each candidate point $\vx'_i \in \mathcal{X}'$, every local model $GP_{t-1}^{(m)}$ independently computes its posterior predictive mean $\mu_{t-1,m}(\vx'_i)$ and variance $\sigma^2_{t-1,m}(\vx'_i)$ as follows:
\begin{align}
\mu_{t-1,m}(\vx'_i) &=  \tilde{\mu}_{t-1}^{(m)}(\vx'_i) +  \vk_{t-1}^{(m)}(\vx'_i)^\top \left(  \mK_{t-1}^{(m)} + \ddot{\sigma}_{t-1,m}^2 \ \mI \right)^{-1}  \left( \vy_{t-1}^{(m)} - \tilde{\vmu}_{t-1}^{(m)}  \right) \\
\sigma_{t-1,m}^2(\vx'_i) &= k_{t-1}^{(m)}(\vx'_i, \vx'_i) -  \vk_{t-1}^{(m)}(\vx'_i)^\top  \left(  \mK_{t-1}^{(m)} + \ddot{\sigma}_{t-1,m}^2 \ \mI  \right)^{-1} \vk_{t-1}^{(m)}(\vx'_i) 
\label{eq:eq_c5_2}
\end{align}
where $\tilde{\mu}_{t-1}^{(m)}$ denotes the mean function, $k_{t-1}^{(m)}$ denotes the covariance function, and $\vk_{t-1}^{(m)}$ denotes the covariance vector between $\vx'_i$ and $\mX_{t-1}^{(m)}$.

To mitigate variance overestimation, the predictive variance of each local GP model is adjusted using a calibration ratio $\alpha_{t-1,m}(\vx'_i)$, yielding the calibrated variance $\hat{\sigma}^2_{t-1,m}(\vx'_i)$:
\begin{align}
\hat{\sigma}_{ t-1,m}^2(\vx'_i) =  \alpha_{ t-1,m} (\vx'_i) \  \cdot \ \sigma_{ t-1,m}^2(\vx'_i) \label{eq:eq_c6_cc_ratio}
\end{align}
where $0 < \alpha_{t-1,m}(\vx'_i) \leqslant 1$ and the calibration ratio is computed as:
\begin{align}
 \alpha_{t-1,m}(\vx'_i) = \exp \left( -  \  \frac{e-1}{e}  \  \frac{1}{2} \log \left( 1 + \frac{ \sigma_{t-1,m}^2(\vx'_i) } { \ddot{\sigma}_{t-1,m}^2 }  \right)  \right)  
\end{align}
as detailed in \cite{Ong2026}.

The aggregated predictive mean $\mu_{t-1}(\vx'_i)$ and variance $\sigma^2_{t-1}(\vx'_i)$ are then computed as:
\begin{align}
\mu_{t-1}(\vx'_i) &= \sigma_{t-1}^2(\vx'_i)  \sum_{m=1}^M \left( w_{t-1,m}(\vx'_i) \  \mu_{t-1,m}(\vx'_i) \ \frac{1}{ \hat{\sigma}_{t-1,m}^2(\vx'_i)} \right)  \label{eq:eq_c6_agg_mu} \\
\sigma_{t-1}^2(\vx'_i) &= \left( \sum_{m=1}^M \left( w_{t-1,m}(\vx'_i) \  \frac{1}{ \hat{\sigma}_{t-1,m}^2(\vx'_i)}  \right)  \right)^{-1} \label{eq:eq_c6_agg_sigma}
\end{align}
where $w_{t-1,m}(\vx'_i)\in\mathbb{R}^{+}$ denotes the weight assigned to the $m$-th local GP model. The weights are computed using the entropy-weighted aggregation scheme as described in \cite{Ong2026}.

\paragraph{Selecting query points } \hfill

Let $\mathcal{X}'$ denote the union of the candidate sets generated by all trust regions. At iteration $t$, an acquisition function is used to construct a batch of $B$ query points $\{\vx_{t,1},\ldots,\vx_{t,B}\}$ from $\mathcal{X}'$.
Three acquisition functions are considered: (i) expected improvement (EI), (ii) Gaussian process lower confidence bound (GP-LCB), and (iii) Thompson sampling (TS).

The query batch is constructed greedily. At each selection step, the candidate with the best acquisition value is selected and removed from the candidate set before the next query point is chosen. For EI and TS, candidates with the largest acquisition values are preferred, whereas GP-LCB selects candidates with the smallest acquisition values.

The selected query points are evaluated using the objective function, yielding observations \[ y_{t,j}=f(\vx_{t,j})+\epsilon_{t,j}, \qquad j=1,\ldots,B \] The resulting observations are then used to update the trust regions.

\paragraph{Updating trust regions} \hfill

Let $V^{(q)}$ denote the number of selected query points originating from the candidate set associated with trust region $TR^{(q)}$. A batch is considered successful for $TR^{(q)}$ if $V^{(q)} > 0$ and at least one selected query point produces an observation that improves upon the current best observation associated with that trust region (i.e., the observation at its centre). In this case, the success counter is incremented, the failure counter is reset to zero, and the centre of $TR^{(q)}$ is updated to the best-performing query point.

If $V^{(q)} > 0$ but none of the selected query points improves upon the current best observation, the batch is considered unsuccessful. The failure counter is incremented and the success counter is reset to zero. If $V^{(q)} = 0$, the inactivity counter is incremented by one.

When the success counter exceeds the threshold $\tau_{succ}$, the base side length of the trust region is expanded according to \[ L^{(q)} \leftarrow \min\{L_{max},\,2L^{(q)}\} \] 
Conversely, when the failure counter exceeds $\tau_{fail}$, the side length is reduced according to \[ L^{(q)} \leftarrow \frac{L^{(q)}}{2} \]
After either update, the success and failure counters are reset to zero.

A trust region is discarded if either $ L^{(q)} < L_{min} $ or its inactivity counter exceeds $ \tau_{inac} $.
Importantly, discarding a trust region does not remove the associated local GP model or any previously collected observations. Consequently, all evaluated data remain available throughout the optimisation process, eliminating the information loss encountered in conventional trust-region methods such as TuRBO.

The optimisation procedure continues until the evaluation budget $T$ is exhausted. The final recommendation is \[ \hat{\vx}_T = \argmin_{\vx \in \mathcal{D}_T} y(\vx), \] corresponding to the input associated with the lowest observed objective value.

\section{Experiments}
\label{sec:sec_c5_experiments}

This section presents the results and discussion of empirical evaluations of BO-pro-c-tr.
Three sets of experiments were conducted. The first set used 16 synthetic benchmark functions to evaluate the performance and computational overhead of BO-pro-c-tr with respect to acquisition function and input dimensionality. The second and third sets evaluated the performance and computational overhead of BO-pro-c-tr on two simulation tasks.

For clarity, all experiments followed a common evaluation protocol, described below.

\textbf{BO algorithms}. 
The experiments compared BO-pro-c-tr against 
(i) TuRBO, which employs a collection of independent local GP models associated with trust regions in a one-to-one manner; 
(ii) BO algorithms based on a product-of-experts GP model with uncertainty calibration (BO-pro-c); and 
(iii) BO algorithms based on a single global GP model (BO-glo).
In total, 13 BO algorithm variants were considered: 
\begin{enumerate}[(i)] 
\item six BO-pro-c-tr variants, corresponding to the Cartesian product of two local GP sizes (200 and 400 data points per local GP model) and three acquisition functions, while using the ball-tree assignment method and entropy-weighted aggregation throughout; 
\item three BO-glo variants using the EI, GP-LCB, and TS acquisition functions, respectively; 
\item three BO-pro-c variants using the ball-tree assignment method, entropy-weighted aggregation, 200 data points per local GP model, and the EI, GP-LCB, and TS acquisition functions; and 
\item one TuRBO variant using two independent local GP models. 
\end{enumerate}

Table~\ref{tab:tab_c6_gp_models} summarises the BO algorithms used in the experiments. 
The algorithm names (e.g., \texttt{pro-c-tr-ei-200}) encode the key characteristics of each method. 
The prefixes \texttt{pro}, \texttt{glo}, and \texttt{turbo} denote product-of-experts GP models, a single global GP model, and the TuRBO algorithm, respectively. 
The suffix \texttt{c} indicates the use of uncertainty calibration, whereas \texttt{tr} denotes the one-to-many association between local GP models and trust regions. 
The labels \texttt{ei}, \texttt{lcb}, and \texttt{ts} specify the acquisition function. 
Finally, the trailing number (200 or 400) indicates the maximum number of data points assigned to each local GP model.

\begin{table}
\centering
\caption{ Summary of the BO algorithms used in the experiments for evaluating and comparing BO-pro-c-tr, BO-pro-c, TuRBO, and BO-glo. }
\begin{scriptsize}
\begin{tabular}[t]{llllll}
\toprule
Algorithm & GP model (number of data        & Association     &  Data      &  Weighting    &  Acquisition       \\
	        & points per local GP)            & of GP-TR        &    assignment                &  scheme         &   function   \\
\midrule
glo-ei                           &  single global GP                    &       &       &          &      EI      \\
glo-lcb                           &  single global GP                    &       &       &       &      GP-LCB      \\
glo-ts                           &  single global GP                    &       &       &         &      TS      \\
\vspace{1\baselineskip}\\  
turbo-2			& 2 independent local GPs       & one-to-one  &    &      &     \\
\vspace{1\baselineskip}\\  
pro-c-ei         &  product-of-experts GP (200)      &       &        ball-tree            &   entropy     &      EI    \\
pro-c-lcb        &  product-of-experts GP (200)      &       &      ball-tree              &   entropy     &      GP-LCB    \\
pro-c-ts         &  product-of-experts GP (200)      &       &    ball-tree                &   entropy     &      TS    \\
\vspace{1\baselineskip}\\  
pro-c-tr-ei-200         &  product-of-experts GP (200)      & one-to-many      &     ball-tree           &   entropy     &      EI     \\
pro-c-tr-lcb-200         &  product-of-experts GP (200)      & one-to-many      &     ball-tree           &   entropy     &      GP-LCB  \\
pro-c-tr-ts-200         &  product-of-experts GP (200)      & one-to-many      &     ball-tree           &   entropy     &      TS     \\
\vspace{1\baselineskip}\\  
pro-c-tr-ei-400         &  product-of-experts GP (400)      & one-to-many      &     ball-tree           &   entropy     &      EI     \\
pro-c-tr-lcb-400         &  product-of-experts GP (400)      & one-to-many      &     ball-tree           &   entropy     &      GP-LCB  \\
pro-c-tr-ts-400         &  product-of-experts GP (400)      & one-to-many      &     ball-tree           &   entropy     &      TS     \\
\bottomrule
\label{tab:tab_c6_gp_models}
\end{tabular}
\end{scriptsize}
\end{table}

\textbf{GP modelling}. 
GPyTorch \citep{Gardner2018} was used for GP regression in all experiments. Before model fitting, the input domain was rescaled to $[0,1]^d$, and the function values were standardised. Each GP employed a Matérn-$5/2$ covariance function with automatic relevance determination (ARD) and a constant mean function. GP hyperparameters were estimated by maximising the log marginal likelihood prior to selecting new query points.

\textbf{Experimental protocol}.
For each benchmark function, all 13 BO algorithm variants were evaluated, and each experiment was repeated 10 times using different random seeds. 
The reported results correspond to the mean and standard error across the 10 runs. 
Within each run, all algorithms were initialised using the same set of input points to ensure a fair comparison. 
Batches of query points were constructed greedily by optimising the acquisition function. At each selection step, the candidate with the best acquisition value was selected and removed from the candidate set before the next query point was chosen. The evaluation of an entire batch was counted as a single BO iteration.

\subsection{Experiments on synthetic benchmark functions}

This subsection evaluates BO-pro-c-tr on 16 synthetic benchmark functions and compares its performance against TuRBO, BO-pro-c, and BO-glo. The experiments investigate the effect of (i) acquisition function and (ii) input dimensionality on optimisation performance and computational overhead.

\textbf{Evaluation metrics}.
The optimisation performance of a BO algorithm after an evaluation budget of $T$ was measured using the simple regret $r_T$ \citep{Bubeck2009,Dorard2012,Shahriari2016}, defined as
\[
r_T = \min_{\vx' \in \mathcal{D}_T} y(\vx') - \min_{\vx \in \mathcal{X}} f(\vx),
\]
that is, the difference between the objective value of the best queried point and that of the true global minimiser.

For pairwise comparisons between BO algorithms, simple-regret values obtained from individual runs were analysed using a paired Student's t-test. The resulting $p$-values were used as an indication of the strength of evidence for performance differences between algorithms. Because each experiment was repeated 10 times using identical random seeds across algorithms, paired comparisons could be performed directly. Differences with $p \leq 0.1$ were regarded as statistically significant.

Computational overhead was defined as the time required to select the next batch of query points.
This included fitting the GP surrogate models and evaluating the acquisition functions.

\textbf{Synthetic benchmark functions}. 
Experiments were conducted on 16 synthetic benchmark functions with known global minima, comprising the Ackley, Levy, Rastrigin, and Rosenbrock functions in dimensions $d=10$, $20$, $50$, and $100$. 
Table~\ref{tab:tab_c6_bo_benchmark_fns} summarises the benchmark functions used in the evaluation. 

All objective function evaluations were corrupted with additive Gaussian noise sampled from $\mathcal{N}(0, 0.25)$. 
Each experiment was repeated 10 times using different random seeds. 
Within each run, all BO algorithms used a batch size of $B=d$. A total evaluation budget of 50 BO iterations was allocated for the $10d$ and $20d$ benchmark functions, while 60 BO iterations were allocated for the $50d$ and $100d$ functions.

For computational-overhead measurements, experiments on the $10d$ and $20d$ benchmark functions were performed on a single Apple M2 processor (3.49 GHz, 8 GB memory), whereas experiments on the $50d$ and $100d$ benchmark functions were performed on Google Colab using an Intel Xeon processor (2.20 GHz, 13 GB memory).

\begin{table}
\centering
\caption{Summary information of the 16 synthetic benchmark functions used in the BO experiments for comparing and evaluating the BO-pro-c-tr, BO-pro-c, TuRBO, and BO-glo algorithms.}
\begin{footnotesize}
\begin{tabular}[t]{llllll}
\toprule
Function & Number of &  Input    &  Minimiser   &  Global minimum  &  \\
	      &  variables  & domain  & $\vx_*$       & $f(\vx_*)$ & \\
\midrule

Ackley          &  10          &  $[-10, 10]^{10}$       &   (0, $\dots$, 0)     & 0      \\
Levy          &  10          &  $[-10, 10]^{10}$       &  (1, $\dots$, 1)     & 0      \\
Rastrigin          &  10          &  $[-5.12, 5.12]^{10}$       &  (0, $\dots$, 0)     & 0      \\
Rosenbrock          &  10          &  $[-5, 10]^{10}$       &  (1, $\dots$, 1)      & 0      \\

Ackley          &  20          &  $[-10, 10]^{20}$       &   (0, $\dots$, 0)     & 0      \\
Levy          &  20          &  $[-10, 10]^{20}$       &  (1, $\dots$, 1)     & 0      \\
Rastrigin          &  20          &  $[-5.12, 5.12]^{20}$       &  (0, $\dots$, 0)     & 0      \\
Rosenbrock          &  20          &  $[-5, 10]^{20}$       &  (1, $\dots$, 1)      & 0      \\

Ackley          &  50          &  $[-10, 10]^{50}$       &   (0, $\dots$, 0)     & 0      \\
Levy          &  50          &  $[-10, 10]^{50}$       &  (1, $\dots$, 1)     & 0      \\
Rastrigin          &  50          &  $[-5.12, 5.12]^{50}$       &  (0, $\dots$, 0)     & 0      \\
Rosenbrock          &  50          &  $[-5, 10]^{50}$       &  (1, $\dots$, 1)      & 0      \\

Ackley          &  100          &  $[-10, 10]^{100}$       &   (0, $\dots$, 0)     & 0      \\
Levy          &  100          &  $[-10, 10]^{100}$       &  (1, $\dots$, 1)     & 0      \\
Rastrigin          &  100          &  $[-5.12, 5.12]^{100}$       &  (0, $\dots$, 0)     & 0      \\
Rosenbrock          &  100          &  $[-5, 10]^{100}$       &  (1, $\dots$, 1)      & 0      \\

\bottomrule
\label{tab:tab_c6_bo_benchmark_fns}
\end{tabular}
\end{footnotesize}
\end{table}

\textbf{Presentation of results}.
To provide a comprehensive overview of optimisation performance and computational overhead, the results were analysed using 
(i) performance progress plots, 
(ii) computational-overhead progress plots, 
(iii) performance--overhead trade-off plots, 
(iv) violin plots,
(v) summary-statistics tables, and 
(vi) $p$-value tables.

Performance progress plots show the average accumulated simple regret across 10 runs at each optimisation step. Computational-overhead progress plots show the average optimisation overhead at each BO iteration. Performance--overhead trade-off plots compare the final simple regret against the average computational overhead over the entire optimisation run.
All simple regret values are reported on a logarithmic scale.

Violin plots illustrate the distribution of final simple regrets across 10 runs. The horizontal lines within each violin indicate the mean, median, minimum, and maximum values. The vertical axis represents the final simple regret on a logarithmic scale.

Below each violin plot, the first table reports the mean, geometric mean, and median final simple regret. Shaded cells identify the three best-performing algorithms according to each statistic.

The second table reports $p$-values obtained from paired $t$-tests comparing the final simple regrets of different algorithms. The ordering of rows and columns is identical. An entry indicates that the algorithm corresponding to the column outperformed the algorithm corresponding to the row, whereas an empty cell indicates the opposite. Values with $p \leqslant 0.1$ are highlighted to indicate statistically significant differences.

\begin{figure}
\centering
\includegraphics[width=1.0\textwidth]{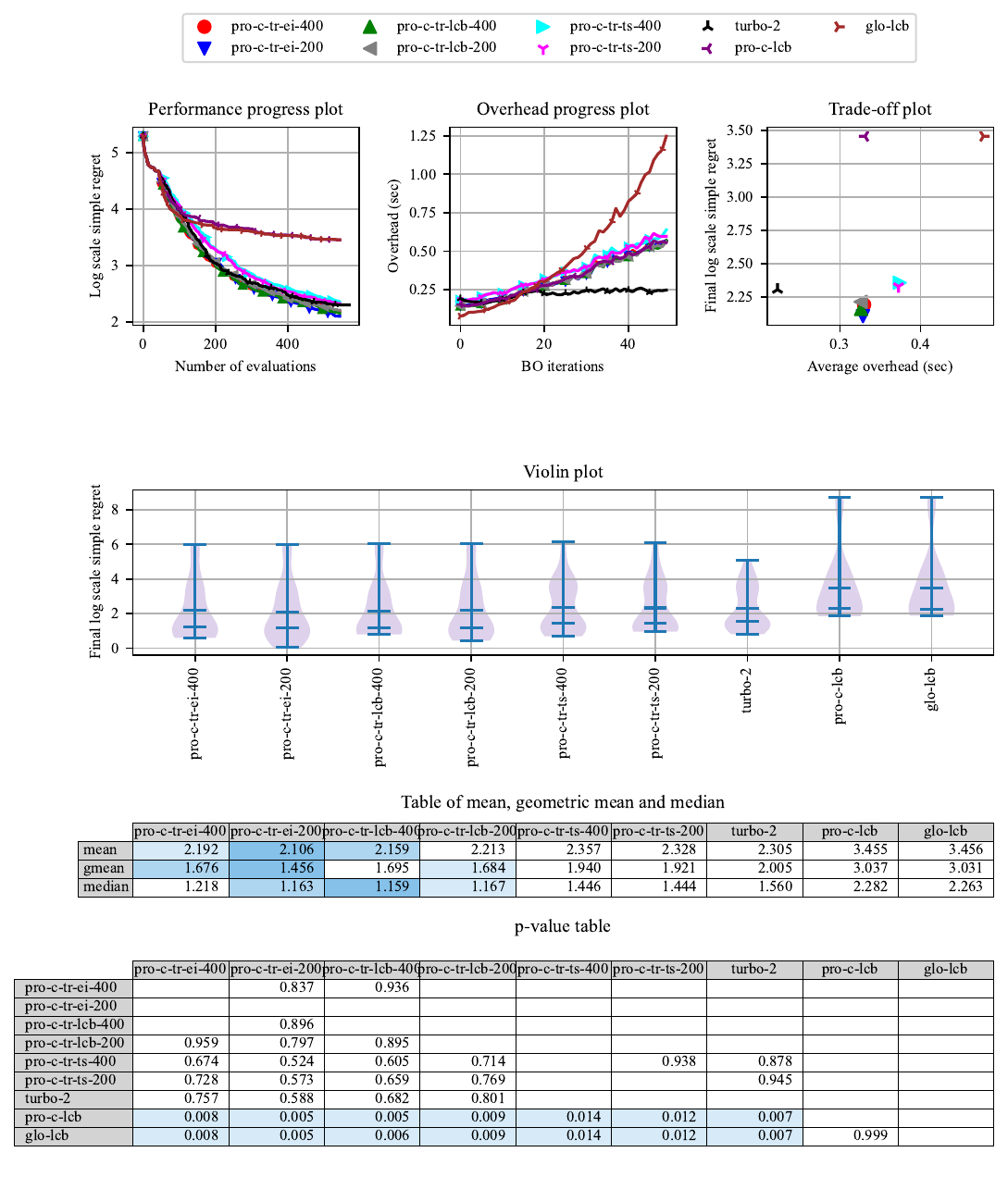}
\caption{
Experiments on four synthetic benchmark functions: Ackley, Levy, Rastrigin, and Rosenbrock, with $d = 10$.
All variants of BO-pro-c-tr-ei and BO-pro-c-tr-lcb algorithms achieved better performance than the other algorithms, while incurring only slightly higher computational overhead than TuRBO. 
}
\label{fig:fig_c6_fn_regret_time_progress_violin_acq_d10}
\end{figure}

\begin{figure}
\centering
\includegraphics[width=1.0\textwidth]{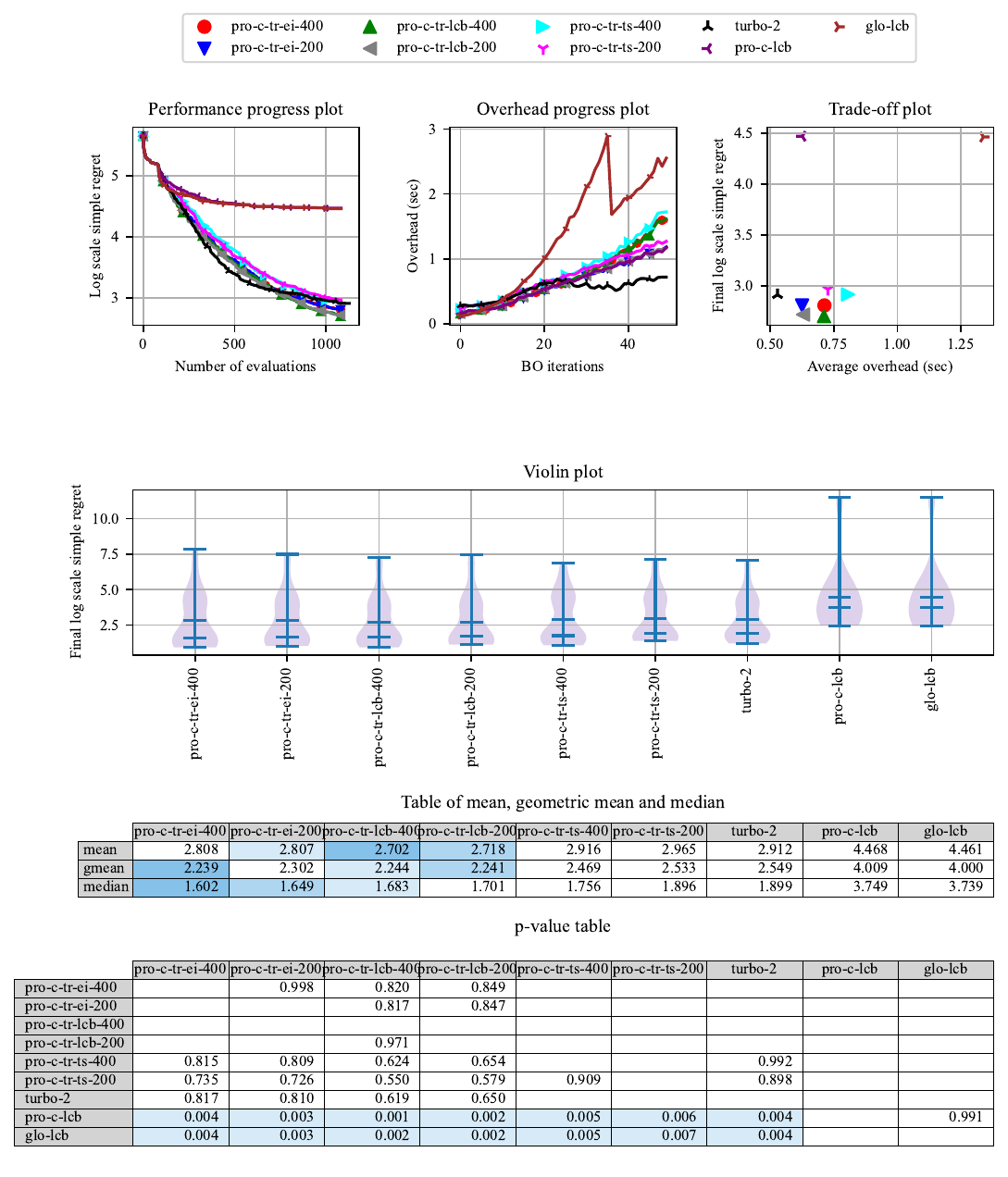}
\caption{  
Experiments on four synthetic benchmark functions: Ackley, Levy, Rastrigin, and Rosenbrock, with $d = 20$.
All variants of BO-pro-c-tr-ei and BO-pro-c-tr-lcb algorithms achieved better performance than the other algorithms, while incurring only slightly higher computational overhead than TuRBO. 
}
\label{fig:fig_c6_fn_regret_time_progress_violin_acq_d20}
\end{figure}

\begin{figure}
\centering
\includegraphics[width=1.0\textwidth]{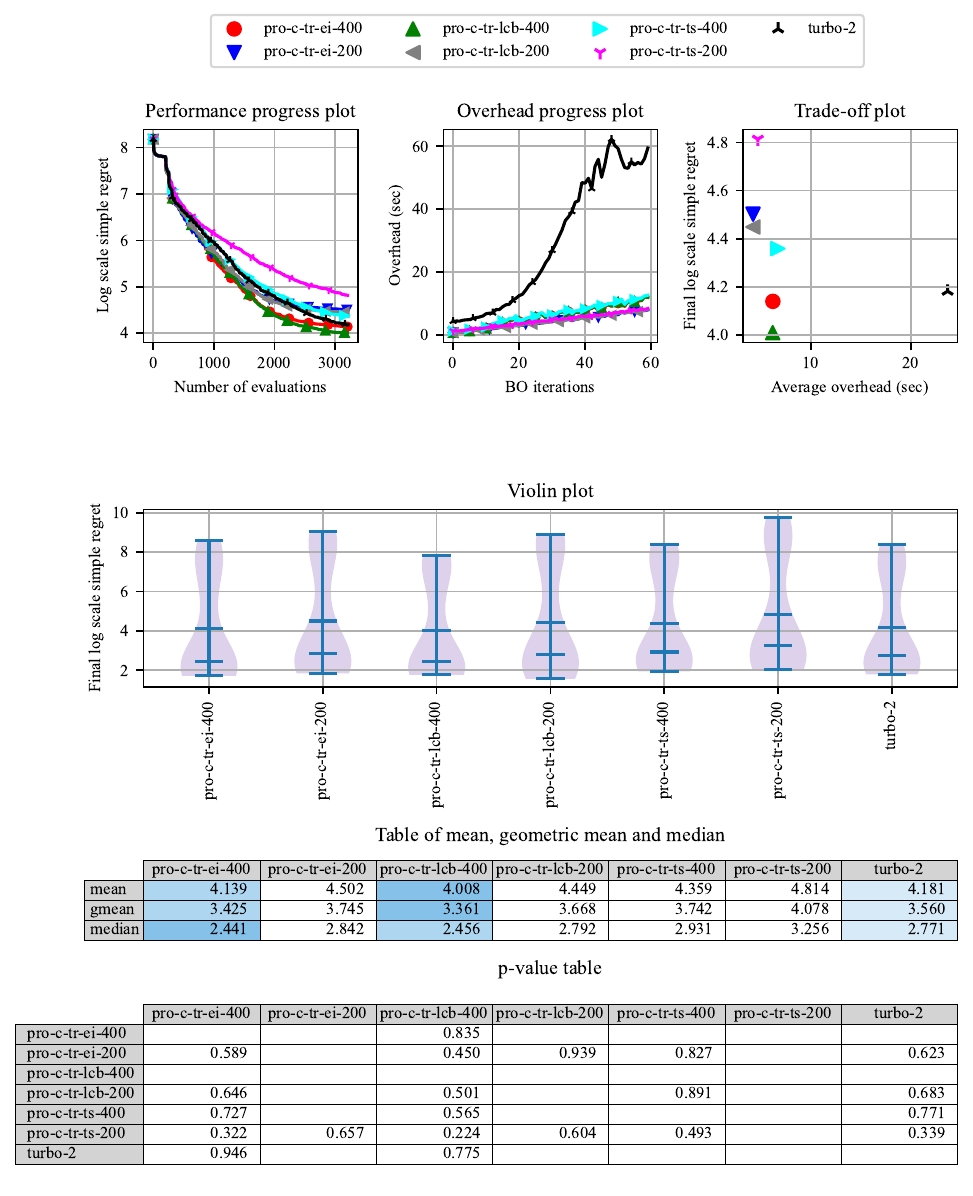}
\caption{  
Experiments on four synthetic benchmark functions: Ackley, Levy, Rastrigin, and Rosenbrock, with $d = 50$.
BO-pro-c-tr-ei and BO-pro-c-tr-lcb algorithms using 400 data points per local GP achieved the best performance while maintaining relatively low computational overheads. 
Note that TuRBO incurred an abrupt increase in computational overhead during the later stages of optimisation because one of the local GP models accumulated a large number of training data points while the associated trust region had not yet met the threshold criteria for discarding the local GP model and trust region. 
}
\label{fig:fig_c6_fn_regret_time_progress_violin_acq_d50}
\end{figure}

\begin{figure}
\centering
\includegraphics[width=1.0\textwidth]{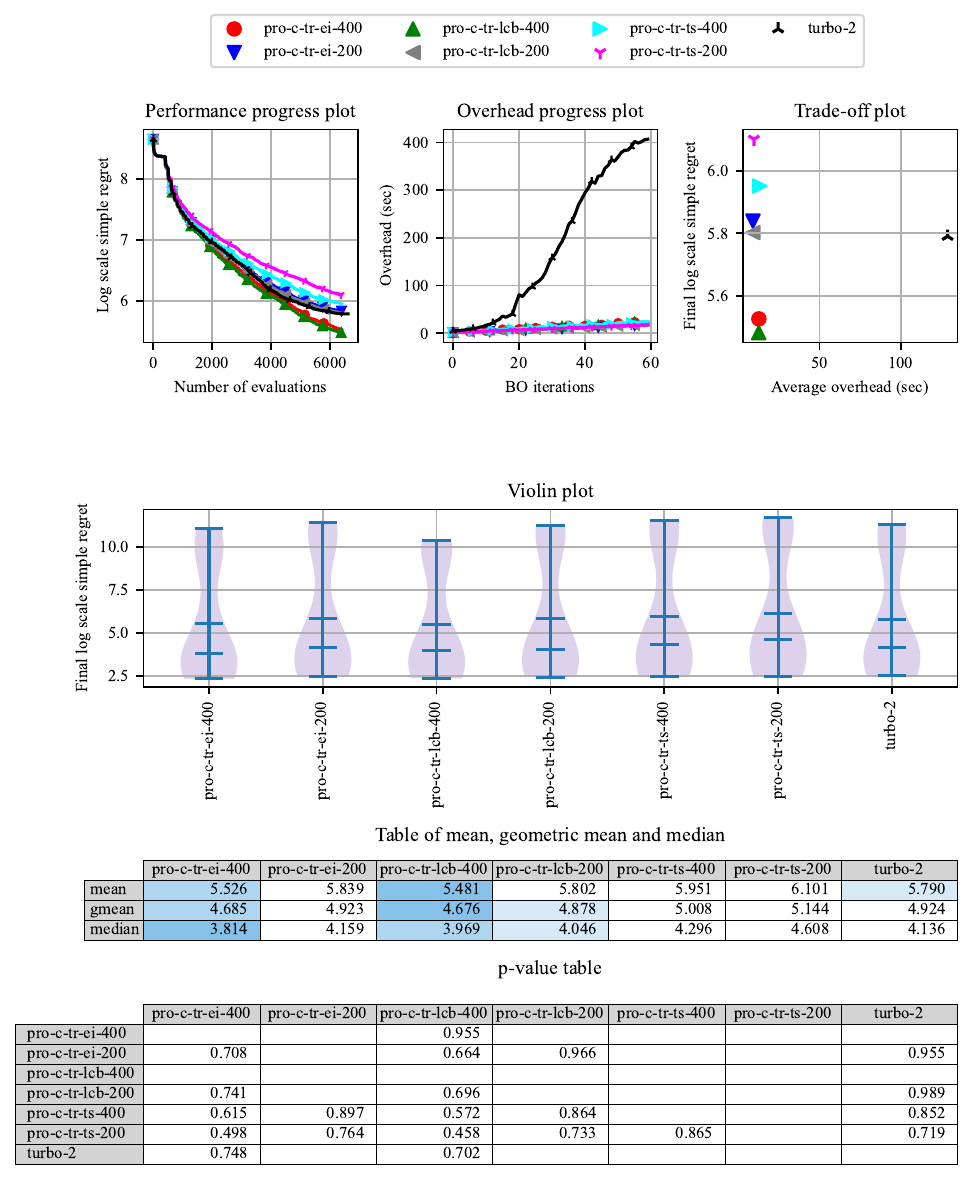}
\caption{  
Experiments on four synthetic benchmark functions: Ackley, Levy, Rastrigin, and Rosenbrock, with $d = 100$.
BO-pro-c-tr-ei and BO-pro-c-tr-lcb algorithms using 400 data points per local GP achieved the best performance while maintaining lower computational overheads. 
Note that TuRBO incurred an abrupt increase in computational overhead during the later stages of optimisation because one of the local GP models accumulated a large number of training data points while the associated trust region had not yet met the threshold criteria for discarding the local GP model and trust region. 
}
\label{fig:fig_c6_fn_regret_time_progress_violin_acq_d100}
\end{figure}

\textbf{Analysis of results}.
Figures~\ref{fig:fig_c6_fn_regret_time_progress_violin_acq_d10}, \ref{fig:fig_c6_fn_regret_time_progress_violin_acq_d20}, \ref{fig:fig_c6_fn_regret_time_progress_violin_acq_d50}, and \ref{fig:fig_c6_fn_regret_time_progress_violin_acq_d100} present performance and computational-overhead analyses as functions of the acquisition function. 
Each figure reports results for four benchmark functions with input dimensionalities $d=10$, $20$, $50$, and $100$, respectively. For each dimensionality, results were averaged across the Ackley, Levy, Rastrigin, and Rosenbrock functions. 
Algorithms were grouped according to acquisition function and the maximum number of data points assigned to each local GP model. 
For brevity, only the GP-LCB variants of BO-pro-c and BO-glo were included.

Overall, all BO-pro-c-tr variants (pro-c-tr) and TuRBO (turbo-2) substantially outperformed BO-pro-c (pro-c) and BO-glo (glo) across all acquisition functions. Many of the observed performance differences were statistically significant. These findings highlight the importance of trust-region methods for high-dimensional Bayesian optimisation. As expected, BO-glo performed poorly in terms of both optimisation performance and computational overhead as dimensionality increased.

For the $10d$ and $20d$ benchmark functions, BO-pro-c-tr consistently achieved better optimisation performance than TuRBO, although it incurred slightly higher computational overhead. This additional overhead arises because BO-pro-c-tr retains all previously evaluated observations, whereas TuRBO periodically discards trust regions and their associated GP models.

For the $50d$ and $100d$ benchmark functions, BO-pro-c-tr not only achieved better optimisation performance than TuRBO, but also maintained a substantially more stable growth in computational overhead throughout optimisation. 
In contrast, TuRBO exhibited abrupt increases in computational overhead during the later stages of optimisation, as shown in Figures~\ref{fig:fig_c6_fn_regret_time_progress_violin_acq_d50} and~\ref{fig:fig_c6_fn_regret_time_progress_violin_acq_d100}.
Inspection of the corresponding optimisation runs revealed that these spikes occurred when one of TuRBO's local GP models accumulated a large number of observations before its associated trust region met the criteria for resetting. As a consequence, the local model temporarily behaved like a global GP, reintroducing the computational bottleneck associated with GP regression. This effect became increasingly pronounced as input dimensionality increased.

Overall, the results suggest that BO-pro-c-tr provides a more effective combination of trust-region optimisation and scalable surrogate modelling than TuRBO. Across all experimental settings considered, BO-pro-c-tr achieved superior optimisation performance while maintaining lower and more predictable computational overhead. The strongest overall performance was obtained when BO-pro-c-tr was combined with the GP-LCB and EI acquisition functions.

\subsection{Robot pushing task}

The second experiment evaluated BO-pro-c-tr and TuRBO on a policy-search task involving the control of two robotic manipulators in a 14-dimensional optimisation problem.

Reinforcement learning plays a central role in robotics, and Bayesian optimisation has been shown to be an effective approach for data-efficient policy search in reinforcement learning settings \citep{Wilson2014,Paul2020,Javier2021}. Compared with exhaustive search methods such as grid search, BO provides a model-based optimisation framework that can efficiently identify high-performing policies within a continuous policy space. Although the tasks considered in this section and the next are less complex than real-world robotic systems, they have been widely used as benchmark problems for evaluating BO algorithms in policy-search applications \citep{Wang2018,Eriksson2019}.

The robot pushing task is implemented using the Box2D physics engine\footnote{\url{https://box2d.org/}}. The task simulates two robot hands pushing two objects across a horizontal surface. The objective is to optimise a controller that enables the robot hands to move both objects to predefined target locations.

The robot controller is parameterised by a 14-dimensional vector specifying the initial positions and orientations of the robot hands, pushing speeds, movement directions, and pushing durations. The lower bounds of the controller parameters are \[ [-5,-5,-10,-10,2,0,-5,-5,-10,-10,2,0,-5,-5] \] and the corresponding upper bounds are \[ [5,5,10,10,30,2\pi,5,5,10,10,30,2\pi,5,5] \]

Within the objective function, the input corresponds to a set of pushing actions performed by the robot hands, while the output represents the resulting distances between the objects and their target locations.

At each optimisation step, the controller parameters are applied to the robot hands, causing the objects to move to new positions. 
The quality of the resulting behaviour is measured using the reward function 
\[ f(\vs) = \sum_{j=1}^{2} \left( \lVert \vs_{gj}-\vs_{ij} \rVert - \lVert \vs_{gj}-\vs_{nj} \rVert \right) \] 
where $\vs_{ij}$ denotes the initial position of object $j$, $\vs_{nj}$ denotes its final position after the pushing action, and $\vs_{gj}$ denotes its target location. 
The objective is to maximise the reward. The maximum reward is achieved when both objects reach their respective target locations.

\textbf{Experimental protocol}.
The experiment compared seven BO algorithms and was repeated 10 times using different random seeds. The reported results correspond to the mean and standard error across the 10 runs. Within each run, all algorithms used a batch size of $B=d=14$ and were allocated a total evaluation budget of 70 BO iterations.

\begin{figure}
\centering
\includegraphics[width=1.0\textwidth]{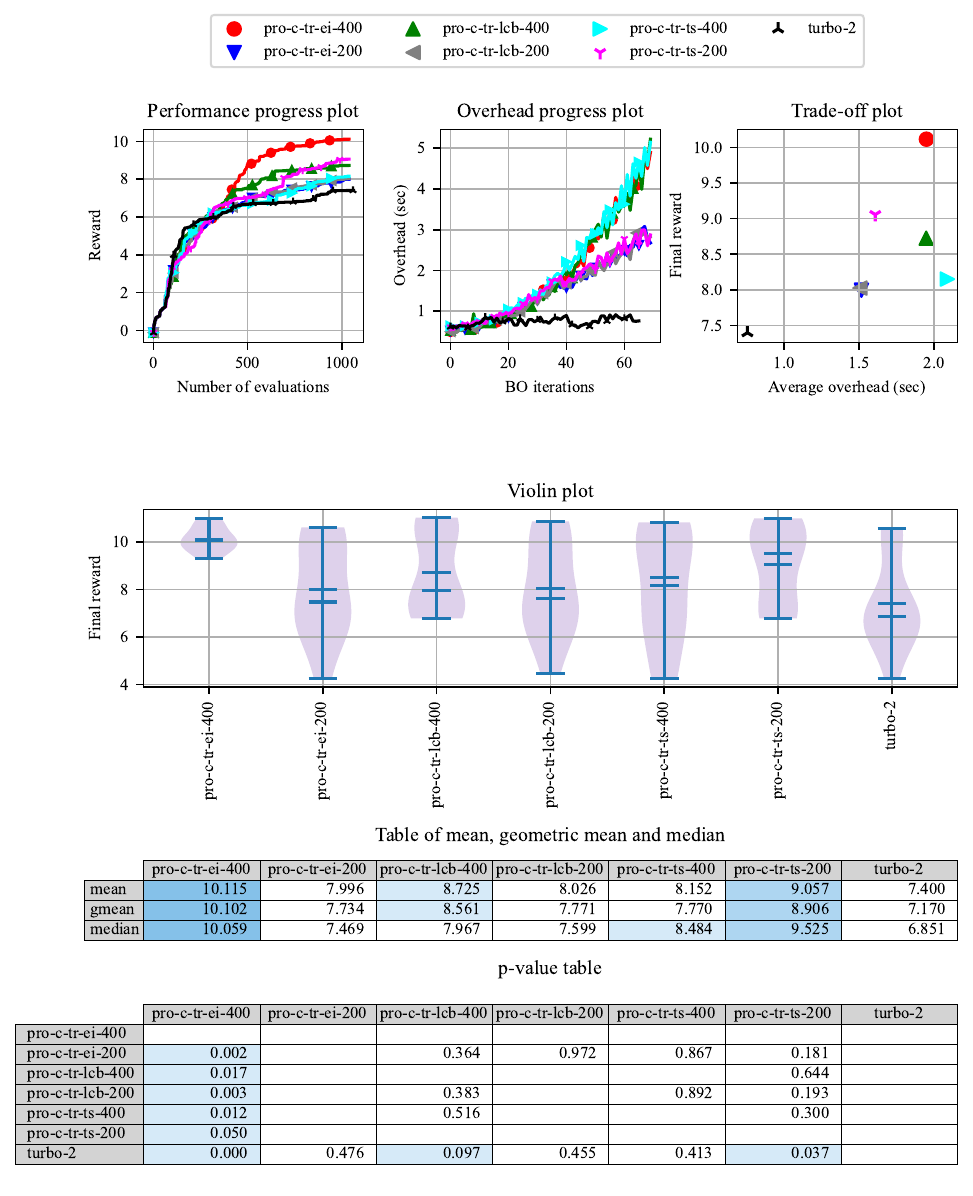}
\caption{ Experiments on the robot pushing task ($d=14$, $B=14$). The figure compares the optimisation performance and computational overhead of BO-pro-c-tr (pro-c-tr) and TuRBO (turbo-2). BO-pro-c-tr using the EI acquisition function with 400 data points per local GP model (pro-c-tr-ei-400) achieved the highest rewards while incurring only a modest increase in computational overhead relative to TuRBO. }
\label{fig:fig_c5_robot_regret_time_progress_violin_acq}
\end{figure}

\textbf{Results}.
Figure~\ref{fig:fig_c5_robot_regret_time_progress_violin_acq} compares the performance and computational overhead of BO-pro-c-tr and TuRBO across three acquisition functions: EI, GP-LCB, and TS. All BO-pro-c-tr variants achieved higher rewards than TuRBO while incurring only a modest increase in computational overhead. Among the evaluated methods, BO-pro-c-tr using the EI acquisition function with 400 data points per local GP model (pro-c-tr-ei-400) achieved the strongest overall performance, attaining the highest rewards and outperforming all competing algorithms by a substantial margin. These results indicate that the combination of uncertainty-calibrated product-of-experts GP modelling and one-to-many trust-region associations provides a more effective policy-search strategy than independent local GP models with one-to-one trust-region associations.

\subsection{Rover trajectory planning task}

The third experiment evaluated BO-pro-c-tr and TuRBO on a 60-dimensional rover trajectory planning problem.

The objective is to identify an optimal trajectory that enables a rover to traverse rough terrain represented by a two-dimensional environment populated with obstacle objects that impede movement. 
The trajectory is constructed by fitting a B-spline through 30 control points in the plane. 
Consequently, the task is to identify a feasible and efficient path through the obstacle field that guides the rover from the starting location $(0.05, 0.05)$ to a position near the target location $(0.95, 0.95)$. 
Collisions with obstacles incur penalties. 
The simulation environment was implemented using the Python path-planning module described by \cite{Wang2018}.

The policy is parameterised by 60 variables defined on $\mathcal{S} \in [0,1]^{60}$. These parameters correspond to the $x$- and $y$-coordinates of 30 control points in the two-dimensional plane. For $\vs \in [0,1]^{60}$, the reward function is defined as 
\[ 
f(\vs) = c(\vs) - 10 \left( \lVert \vs_{1,2} - \vs_s \rVert_1 + \lVert \vs_{59,60} - \vs_g \rVert_1 \right) + 5 
\]
where $c(\vs)$ denotes the trajectory cost function, which assigns a penalty of $-20$ whenever the generated trajectory intersects an obstacle. 
The vectors $\vs_s$ and $\vs_g$ denote the desired start and goal locations, respectively. 
The first two dimensions, $\vs_{1,2}$, and the final two dimensions, $\vs_{59,60}$, represent the start and end points of the generated trajectory. 
Consequently, the reward function penalises trajectories whose start and end locations deviate substantially from the desired positions. The objective is to maximise the reward function.

\textbf{Experimental protocol}.
The experiment compared seven BO algorithms and was repeated 10 times using different random seeds. The reported results correspond to the mean and standard error across the 10 runs. Within each run, all algorithms used a batch size of $B=d=60$ and were allocated a total evaluation budget of 60 BO iterations.

\begin{figure}
\centering
\includegraphics[width=1.0\textwidth]{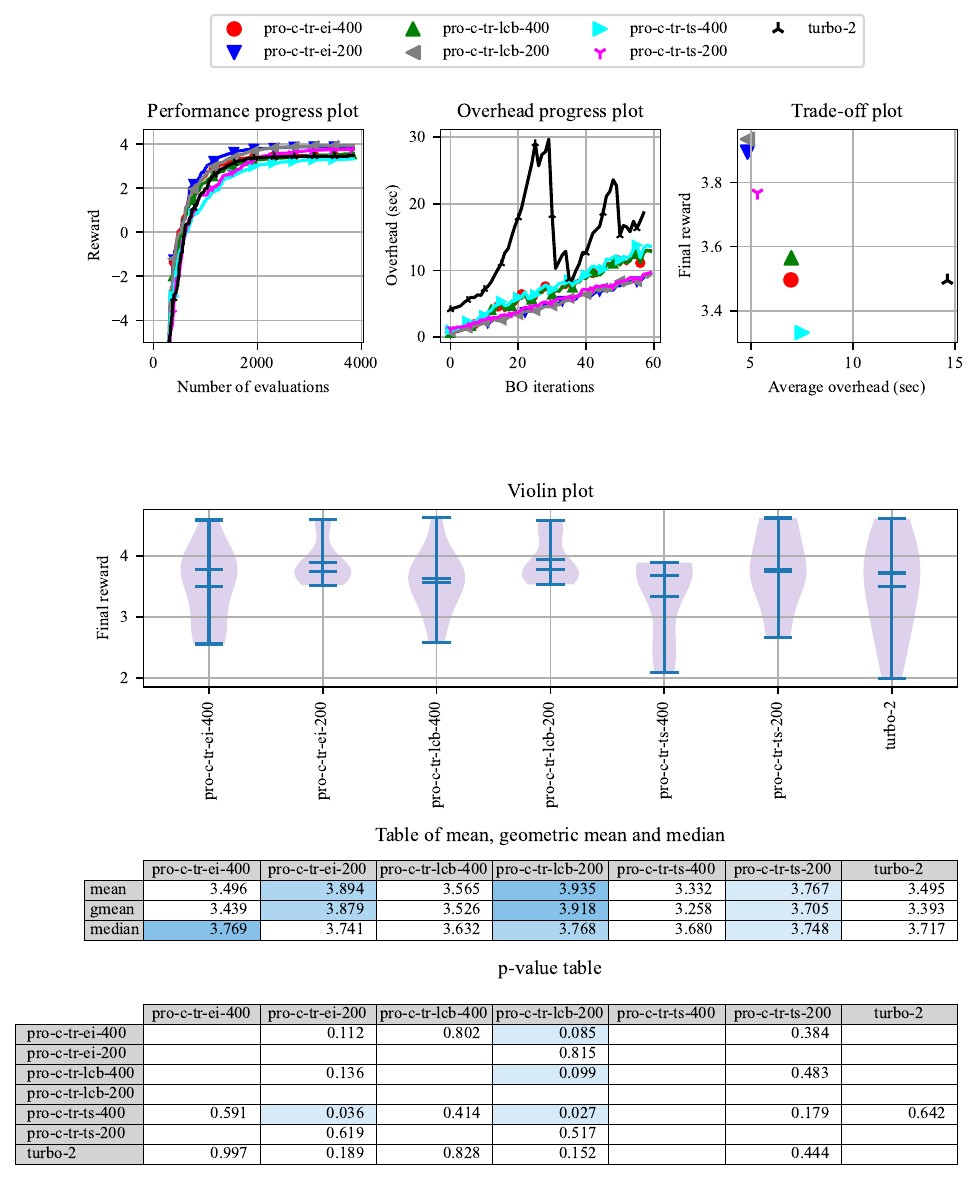}
\caption{ Experiments on the rover trajectory planning task ($d=60$, $B=60$). The figure compares the optimisation performance and computational overhead of BO-pro-c-tr (pro-c-tr) and TuRBO (turbo-2). BO-pro-c-tr using the GP-LCB acquisition function with 200 data points per local GP model (pro-c-tr-lcb-200) achieved the highest rewards, while all BO-pro-c-tr variants maintained lower computational overhead than TuRBO. }
\label{fig:fig_c5_lunar_regret_time_progress_violin_acq}
\end{figure}

\textbf{Results}.
Figure~\ref{fig:fig_c5_lunar_regret_time_progress_violin_acq} compares the performance and computational overhead of BO-pro-c-tr and TuRBO across three acquisition functions: EI, GP-LCB, and TS. 
Among the evaluated methods, BO-pro-c-tr using the GP-LCB acquisition function (pro-c-tr-lcb-200) achieved the highest rewards. The remaining BO-pro-c-tr variants also performed competitively with TuRBO, achieving comparable or better optimisation performance across the repeated runs. 
In terms of computational efficiency, all BO-pro-c-tr variants incurred lower computational overhead than TuRBO. 
These results indicate that the combination of uncertainty-calibrated product-of-experts GP modelling and one-to-many trust-region associations provides an effective and scalable approach for high-dimensional trajectory-planning tasks.


\section{Discussion and conclusion}
\label{sec:sec_s5_conclusion}

Table~\ref{tab:tab_c5_regret_all} reports the average log-scale simple regrets achieved by the BO-pro-c-tr-ei-400, BO-pro-c-tr-lcb-400, BO-pro-c-tr-ts-400, and TuRBO algorithms across 16 synthetic benchmark functions. The corresponding computational overheads are reported in Table~\ref{tab:tab_c5_overhead_all}.

Overall, the experimental results suggest that BO-pro-c-tr provides an effective combination of uncertainty-calibrated product-of-experts GP modelling and trust-region optimisation. Unlike collections of independent local GP models, the GP-pro-c model enables information sharing between local experts and can therefore capture global spatial correlations while preserving local modelling capabilities. The introduction of trust regions further improves scalability by restricting candidate generation to promising regions of the search space. Taken together, these design choices appear to provide advantages over TuRBO, which employs independent local GP models in conjunction with trust regions.

The experiments indicate that BO-pro-c-tr performs particularly well when combined with the ball-tree assignment method, entropy-weighted aggregation, and the GP-LCB acquisition function. These findings are broadly consistent with the results reported for BO-pro-c in \cite{Ong2026_2}.

Compared with TuRBO, BO-pro-c-tr achieved a 2.6\% reduction in average simple regret while reducing computational overhead by 34.9\%. These improvements may be attributed to several factors. 
First, BO-pro-c-tr retains all previously evaluated observations, avoiding the information loss associated with the trust-region reset mechanism used by TuRBO. 
Second, the one-to-many association between local GP models and trust regions provides greater flexibility during optimisation and may reduce the likelihood of the search becoming trapped in local regions of the input space. 
Third, the ball-tree partitioning strategy maintains balanced data allocations across local GP models, resulting in computational overhead that grows more gradually as dimensionality increases.

The empirical results also suggest that uncertainty calibration remains beneficial when product-of-experts GP models are integrated with trust-region methods. Across the benchmark functions and simulation tasks considered in this study, BO-pro-c-tr using GP-LCB generally achieved the strongest overall performance, highlighting the importance of accurate uncertainty estimates when acquisition functions rely directly on predictive confidence intervals.

\begin{table}
\centering
\caption{Average log-scale simple regrets achieved by BO-pro-c-tr-ei-400, BO-pro-c-tr-lcb-400, BO-pro-c-tr-ts-400, and TuRBO across 16 synthetic benchmark functions with input dimensionalities of 10, 20, 50, and 100.}
\begin{scriptsize}
\begin{tabular}[t]{lccccl}
\toprule
Function & BO-pro-c-tr-ei-400   &  BO-pro-c-tr-lcb-400    &  BO-pro-c-tr-ts-400    &  TuRBO   &  \\
\midrule

$10d$ Ackley             &   $0.988 \pm 0.183$    &   $0.987 \pm 0.115$    &   $1.239 \pm 0.253$    &   $1.448 \pm 0.109$   \\
$10d$ Levy                &   $0.992 \pm 0.246$    &   $1.030 \pm 0.151$    &   $1.243 \pm 0.443$    &   $1.243 \pm 0.242$   \\
$10d$ Rastrigin         &   $3.128 \pm 0.625$    &   $2.999 \pm 0.326$    &   $3.519 \pm 0.217$    &   $3.337 \pm 0.413$   \\
$10d$ Rosenbrock      &   $5.863 \pm 0.151$    &   $5.815 \pm 0.208$    &   $5.032 \pm 0.909$    &   $4.524 \pm 0.401$   \\

$20d$ Ackley               &   $1.389 \pm 0.181$    &   $1.499 \pm 0.144$    &   $1.573 \pm 0.110$    &   $1.734 \pm 0.107$    \\
$20d$ Levy                 &   $1.386 \pm 0.408$    &   $1.397 \pm 0.268$    &   $1.664 \pm 0.410$    &   $1.775 \pm 0.288$    \\
$20d$ Rastrigin           &  $4.309 \pm 0.214$    &   $3.950 \pm 0.214$    &   $4.456 \pm 0.194$    &   $4.125 \pm 0.163$    \\
$20d$ Rosenbrock      &   $7.274 \pm 0.607$    &   $6.902 \pm 0.280$    &   $6.425 \pm 0.341$    &   $6.578 \pm 0.590$    \\

$50d$ Ackley             &   $1.932 \pm 0.102$    &   $1.948 \pm 0.079$    &   $2.147 \pm 0.113$    &   $2.082 \pm 0.088$    \\
$50d$ Levy                &   $2.394 \pm 0.323$    &   $2.363 \pm 0.336$    &   $2.837 \pm 0.280$    &   $2.618 \pm 0.377$   \\
$50d$ Rastrigin          &   $5.160 \pm 0.097$    &   $4.931 \pm 0.027$    &   $5.243 \pm 0.119$    &   $4.887 \pm 0.049$    \\
$50d$ Rosenbrock      &   $7.785 \pm 0.535$    &   $7.438 \pm 0.276$    &   $7.826 \pm 0.277$    &   $7.632 \pm 0.392$    \\

$100d$ Ackley             &   $2.483 \pm 0.093$    &   $2.443 \pm 0.065$    &   $2.543 \pm 0.034$    &   $2.597 \pm 0.049$    \\
$100d$ Levy                &   $3.781 \pm 0.105$    &   $3.912 \pm 0.191$    &   $4.208 \pm 0.183$    &   $4.088 \pm 0.153$    \\
$100d$ Rastrigin         &   $6.095 \pm 0.064$    &   $6.069 \pm 0.047$    &   $6.189 \pm 0.132$    &   $5.988 \pm 0.087$    \\
$100d$ Rosenbrock    &   $10.143 \pm 0.496$    &   $9.914 \pm 0.290$    &   $11.032 \pm 0.370$    &   $10.627 \pm 0.541$    \\

\bottomrule
Average $\downarrow$             &   4.069      &   \textbf{3.975}       &  4.199          &  4.080   \\
Median $\downarrow$             &  \textbf{3.454}       &   \textbf{3.455}       &   3.863         &  3.713   \\
\bottomrule
\label{tab:tab_c5_regret_all}
\end{tabular}
\end{scriptsize}
\end{table}

\begin{table}
\centering
\caption{Average computational overhead per optimisation iteration (seconds) incurred by BO-pro-c-tr-ei-400, BO-pro-c-tr-lcb-400, BO-pro-c-tr-ts-400, and TuRBO across 16 synthetic benchmark functions with input dimensionalities of 10, 20, 50, and 100.}
\begin{scriptsize}
\begin{tabular}[t]{lccccl}
\toprule
Function & BO-pro-c-tr-ei-400   &  BO-pro-c-tr-lcb-400    &  BO-pro-c-tr-ts-400    &  TuRBO   &  \\
\midrule

$10d$ Ackley             &   $0.333 \pm 0.014$    &   $0.330 \pm 0.013$    &   $0.374 \pm 0.015$    &   $0.185 \pm 0.017$   \\
$10d$ Levy                &   $0.329 \pm 0.021$    &   $0.322 \pm 0.020$    &   $0.375 \pm 0.021$    &   $0.238 \pm 0.026$   \\
$10d$ Rastrigin          &   $0.330 \pm 0.012$    &   $0.327 \pm 0.011$    &   $0.377 \pm 0.010$    &   $0.195 \pm 0.023$   \\
$10d$ Rosenbrock     &   $0.325 \pm 0.014$    &   $0.327 \pm 0.021$    &   $0.368 \pm 0.025$    &   $0.348 \pm 0.022$    \\

$20d$ Ackley               &   $0.721 \pm 0.028$    &   $0.716 \pm 0.027$    &   $0.808 \pm 0.021$    &   $0.443 \pm 0.036$   \\
$20d$ Levy                  &   $0.694 \pm 0.024$    &   $0.694 \pm 0.029$    &   $0.791 \pm 0.029$    &   $0.586 \pm 0.123$   \\
$20d$ Rastrigin           &   $0.731 \pm 0.022$    &   $0.729 \pm 0.026$    &   $0.821 \pm 0.029$    &   $0.489 \pm 0.033$   \\
$20d$ Rosenbrock      &  $0.695 \pm 0.035$    &   $0.698 \pm 0.024$    &   $0.798 \pm 0.043$    &   $0.753 \pm 0.077$    \\

$50d$ Ackley              &   $6.457 \pm 0.885$    &   $6.425 \pm 0.866$    &   $7.000 \pm 0.898$    &   $9.384 \pm 1.832$   \\
$50d$ Levy                 &   $6.193 \pm 0.929$    &   $6.155 \pm 0.900$    &   $6.654 \pm 0.993$    &   $25.053 \pm 12.424$  \\
$50d$ Rastrigin           &   $6.659 \pm 0.597$    &   $6.702 \pm 0.621$    &   $7.163 \pm 0.698$    &   $23.051 \pm 5.032$   \\
$50d$ Rosenbrock       &   $5.867 \pm 1.110$    &   $5.861 \pm 1.073$    &   $6.338 \pm 1.152$    &   $36.723 \pm 13.991$   \\

$100d$ Ackley              &   $11.742 \pm 1.832$    &   $11.615 \pm 1.807$    &   $12.269 \pm 1.880$    &   $20.672 \pm 7.460$   \\
$100d$ Levy                &   $12.472 \pm 1.338$    &   $12.373 \pm 1.355$    &   $12.985 \pm 1.377$    &   $199.620 \pm 50.598$   \\
$100d$ Rastrigin          &   $12.046 \pm 1.177$    &   $11.948 \pm 1.178$    &   $12.681 \pm 1.269$    &   $92.329 \pm 22.975$   \\
$100d$ Rosenbrock     &   $13.534 \pm 2.151$    &   $13.466 \pm 2.211$    &   $14.166 \pm 2.298$    &   $177.535 \pm 34.233$   \\
\bottomrule
Average $\downarrow$            &  4.945       &   \textbf{4.918}       &   5.248         &  36.725   \\
Median $\downarrow$             &   3.299      &    \textbf{3.295}      &   3.580         &   5.069  \\
\bottomrule
\label{tab:tab_c5_overhead_all}
\end{tabular}
\end{scriptsize}
\end{table}

The experiments reported in this work were restricted to optimisation problems with dimensionalities up to 100 because of computational-resource constraints. Although the results consistently favoured BO-pro-c-tr across the considered settings, additional experiments on problems with dimensionalities greater than 100 would provide stronger evidence regarding the scalability of the proposed approach.

The present study focused exclusively on single-objective, single-task Bayesian optimisation. Many important BO settings remain unexplored, including multi-objective BO \citep{Lobato2016,Belakaria2019}, constrained BO \citep{Lobato2016-c}, and multi-fidelity BO \citep{Kandasamy2017,Takeno2020,Irshad2024}. Given its ability to model large-scale datasets using collections of local GP experts, the GP-pro-c framework may provide a useful foundation for scalable surrogate modelling in these settings. Investigating extensions of BO-pro-c-tr to these optimisation paradigms therefore represents a promising direction for future research.

It would also be valuable to evaluate BO-pro-c-tr on real-world scientific optimisation problems, including experimental sequence design and design-of-experiments tasks. Potential application domains include drug discovery \citep{Colliandre2024}, protein design \citep{Hie2022}, cultured-meat media design \citep{Cosenza2023}, and materials discovery \citep{AnhTran2020}, where optimisation is often expensive and data-efficient search strategies are particularly important.

In summary, this paper introduced BO-pro-c-tr, a Bayesian optimisation algorithm that combines uncertainty-calibrated product-of-experts GP models with a redesigned trust-region mechanism. Empirical evaluations on synthetic benchmark functions and policy-search tasks suggest that BO-pro-c-tr can achieve competitive optimisation performance while maintaining substantially lower computational overhead than TuRBO in higher-dimensional settings. These findings indicate that uncertainty-calibrated product-of-experts GP models provide a promising foundation for scalable Bayesian optimisation.

\bibliographystyle{unsrtnat}
\bibliography{references}  

\end{document}